\documentclass[conference]{IEEEtran}
\usepackage{times}
\usepackage[numbers]{natbib}
\usepackage{multicol}
\usepackage[bookmarks=true]{hyperref}
\usepackage{graphics} 
\usepackage{epsfig} 
\usepackage{times} 
\usepackage{amsmath} 
\usepackage{amssymb}  
\usepackage{amsthm}
\usepackage{hyperref}
\usepackage{multirow}
\usepackage{caption,setspace}
\usepackage{booktabs}
\usepackage{arydshln}
\usepackage{tablefootnote}
\usepackage{csquotes}
\usepackage{footnote}
\makesavenoteenv{tabular}
\makesavenoteenv{table}
\usepackage[]{threeparttable}
\usepackage{breqn}
\usepackage{subcaption}
\usepackage{algorithm, algpseudocode}
\usepackage{float}
\usepackage[dvipsnames]{xcolor}
\let\labelindent\relax
\usepackage{enumitem}
\usepackage{tikz}

\usepackage{graphicx}  
\usepackage{subcaption}  

\DeclareMathOperator*{\argmax}{arg\,max}
\DeclareMathOperator*{\argmin}{arg\,min}
\newtheorem{definition}{Definition}

\newtheorem{assumption}{Assumption}

\begin{document}

\title{Structured Imitation Learning of Interactive Policies through Inverse Games}

\author{%
  \authorblockN{%
    Max M. Sun, Todd Murphey %
  }%
  \authorblockA{%
    Center for Robotics and Biosystems, Northwestern University, Evanston, IL 60208 \\
    Email: msun@u.northwestern.edu \\
    Project website: \textcolor{blue}{\url{https://murpheylab.github.io/inverse-mixed-strategy}}
  }%
}

\maketitle
\allowdisplaybreaks

\begin{abstract}
Generative model-based imitation learning methods have recently achieved strong results in learning high-complexity motor skills from human demonstrations. However, imitation learning of interactive policies that coordinate with humans in shared spaces without explicit communication remains challenging, due to the significantly higher behavioral complexity in multi-agent interactions compared to non-interactive tasks. In this work, we introduce a structured imitation learning framework for interactive policies by combining generative single-agent policy learning with a flexible yet expressive game-theoretic structure. Our method explicitly separates learning into two steps: first, we learn individual behavioral patterns from multi-agent demonstrations using standard imitation learning; then, we structurally learn inter-agent dependencies by solving an inverse game problem. Preliminary results in a synthetic 5-agent social navigation task show that our method significantly improves non-interactive policies and performs comparably to the ground truth interactive policy using only 50 demonstrations. These results highlight the potential of structured imitation learning in interactive settings.
\end{abstract}

\IEEEpeerreviewmaketitle

\section{Introduction}

Advances in generative models have significantly increased the capabilities of imitation learning methods~\cite{chi_diffusion_2024,fu_mobile_2024}—motor skill learning paradigms that generate action policies for robots by capturing the statistical behavioral patterns in human demonstrations—enabling high-complexity tasks that are challenging for conventional model-based methods, such as dexterous manipulation~\cite{wang_temporal_2023,ze_3d_2024}, autonomous driving~\cite{pan_agile_2018,weaver_betail_2024}, and agile locomotion~\cite{bin_peng_learning_2020,fu_humanplus_2024}. However, most existing generative model-based imitation learning methods focus on tasks in non-interactive environments, while real-world deployment requires robots to coordinate actions with humans in shared spaces without explicit communication, such as avoiding collisions during navigation~\cite{perez-darpino_fast_2015,trautman_robot_2015}, coordinating manipulation on the same object~\cite{chisari_learning_2024,christen_learning_2023}, and expressing emotional behaviors on robotic characters~\cite{christen_autonomous_2025}. Since each agent’s action influences all others, such environments require learning interactive policies that not only plan for the robot but also anticipate other agents’ intents and actions. Imitation learning of such policies from multi-agent demonstrations remains an open challenge.

\begin{figure}[t!]
    \centering
    \includegraphics[width=0.49\textwidth]{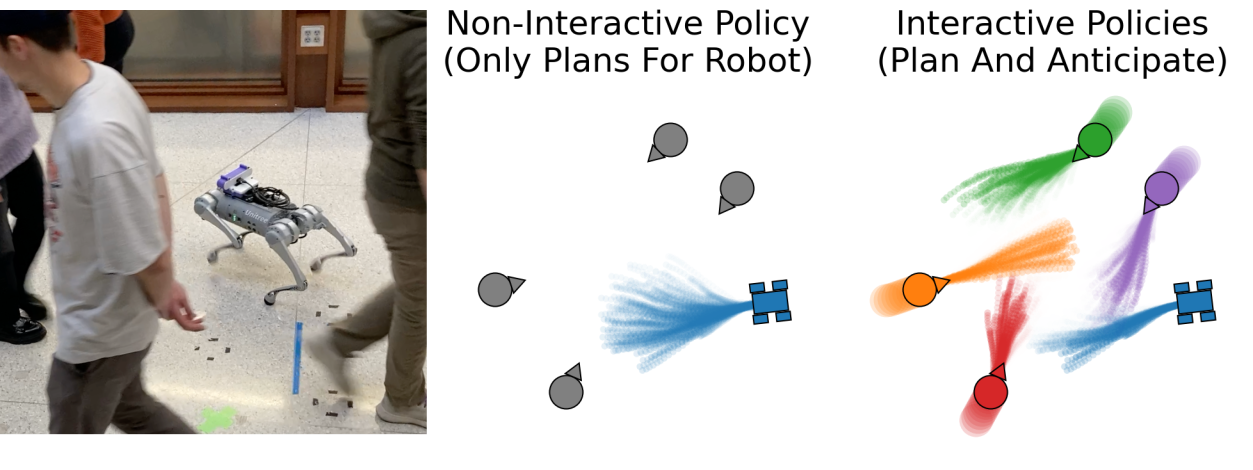}
    \vspace{-2em}
    \caption{For example, social navigation requires the robot to not only plan for itself but also anticipate the actions of surrounding humans to effectively coordinate with them.}
    \label{fig:socnav}
    \vspace{-1em}
\end{figure}

Compared to single-agent policies in non-interactive environments, multi-agent interactions exhibit greater behavioral complexity, making learning difficult~\cite{bobu_aligning_2024}. This complexity arises not just from the number of agents, but also the intertwined nature of decision-making. Each agent's action simultaneously influences all others, introducing extra inter-agent dependencies. Furthermore, each agent’s actions are governed by both individual intents (e.g., reaching a goal) and collective intents shared with the group (e.g., avoiding collisions). Yet, the influence of these intents in demonstrations is subtle, making it especially difficult for learning methods to distinguish and capture behaviors driven by different intents.

\begin{figure*}[t!]
    \centering
    \includegraphics[width=0.95\textwidth]{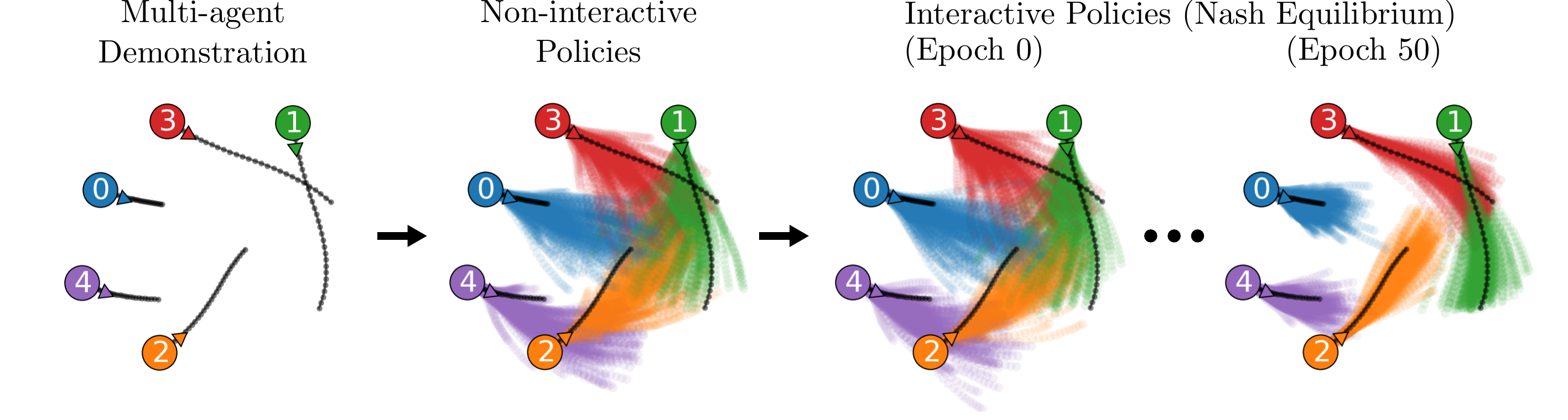}
    \caption{Overview of the structured imitation learning framework. Given a multi-agent demonstration dataset (dark lines indicate demonstrated actions), we first learn the non-interactive policies using standard single-agent imitation learning methods based on generative models. The interactive policies are the Nash equilibrium of a game-theoretic optimization problem based on the non-interactive policies. The cost function of the game-theoretic problem is modeled as a neural network and optimized based on the MLE formula (\ref{eq:game_mle}).}
    \label{fig:overview}
    \vspace{-1em}
\end{figure*}

Instead of relying solely on increasing dataset size to address this complexity, we propose combining generative model-based imitation learning with structured interaction models compatible with generative paradigms~\cite{sun_inverse_2025}. Specifically, our method separates the learning of interactive policies into two stages. First, we leverage generative models to capture individual, non-interactive behavioral patterns from multi-agent data as a standard single-agent imitation learning problem. Then, we model inter-agent dependencies as a game-theoretic optimization problem, whose solution updates the non-interactive policies to incorporate interactions. Importantly, the game structure can be learned from the demonstration data as a neural network, preserving the behavioral model’s expressiveness while improving its ability to capture interactive patterns. We show preliminary results in a 5-agent synthetic social navigation benchmark, where our method significantly improves the non-interactive policy and performs comparably with the ground truth using only 50 demonstrations. These results highlight the potential of structured imitation learning frameworks in interactive environments. Full details of the method can be found in~\cite{sun_inverse_2025}.

\section{Methodology}

\subsection{Problem formulation}

We denote a multi-agent dataset as $\mathcal{D}_{N} = \{d_1, \dots, d_N\}$ containing $N$ expert demonstrations of multi-agent interactions. A demonstration $d_i = \{ \tau_{i,1}, \dots, \tau_{i,M_i}\}$ contains state-action sequences from $M_i$ expert agents. A state-action sequence $\tau_{i,j}={(s_{i,j,1}, a_{i,j,1}), \dots, (s_{i,j,T_i}, a_{i,j,T_i})}$ contains $T_i$ state-actions pairs from a single expert agent. We assume the expert agents are homogeneous and are subject to the same transition dynamics $p(s^\prime | s, a)$. Note that the number of expert agents $M_i$ and the number of time steps $T_i$ contained in one demonstration $d_i$ could vary between demonstrations. 

\begin{definition}[Interactive policy]
    The interactive policy for agent $j$, denoted as $\pi_{\theta}^{(j)}(a_j \vert s_{1\ldots M})$, is a distribution of the agent's actions conditioned on the joint observation of all agents' states.
\end{definition}

We formulate the problem of imitation learning of interactive policies as the following maximum likelihood estimation (MLE) problem:
\begin{align}
    \theta^* = \argmax_{\theta} \sum_{i=1}^{N} \sum_{j=1}^{M_i} \sum_{t=1}^{T_i} \log \pi_{\theta}^{(j)}(a_{i,j,t} \vert s_{i,1\ldots M_i,t}). \label{eq:mle}
\end{align} The MLE formula (\ref{eq:mle}) simultaneously optimizes the interactive policies for all the agents within a demonstration to capture the influences between the agents during decision-making. At the runtime after training, the robot can infer the joint actions of all agents through:
\begin{align}
    \pi_{\theta}(a_{1\ldots M} \vert s_{1\ldots M}) = \prod_{j=1}^{M} \pi_{\theta}^{(j)}(a_j \vert s_{1\ldots M}), \label{ref:factor}
\end{align} which enables the robot to simultaneously plan its own actions while anticipating other agents' actions (see Fig.~\ref{fig:socnav} for an example in social navigation). 

However, learning the interactive policies from multi-agent demonstrations is challenging. Given the same joint observation of all agents' states in a multi-agent demonstration, the interactive policies must produce actions for each agent in a decentralized manner, while remaining close to the joint actions demonstrated by all agents in the dataset. Our goal in this paper is to introduce a structured interaction model into the interactive policy formula and the MLE problem (\ref{eq:mle}) to simplify the learning problem without compromising the effectiveness of the learned policies.

\subsection{Game-theoretic structure for interactive policies}

\begin{figure*}[t!]
    \centering
    \includegraphics[width=0.9\textwidth]{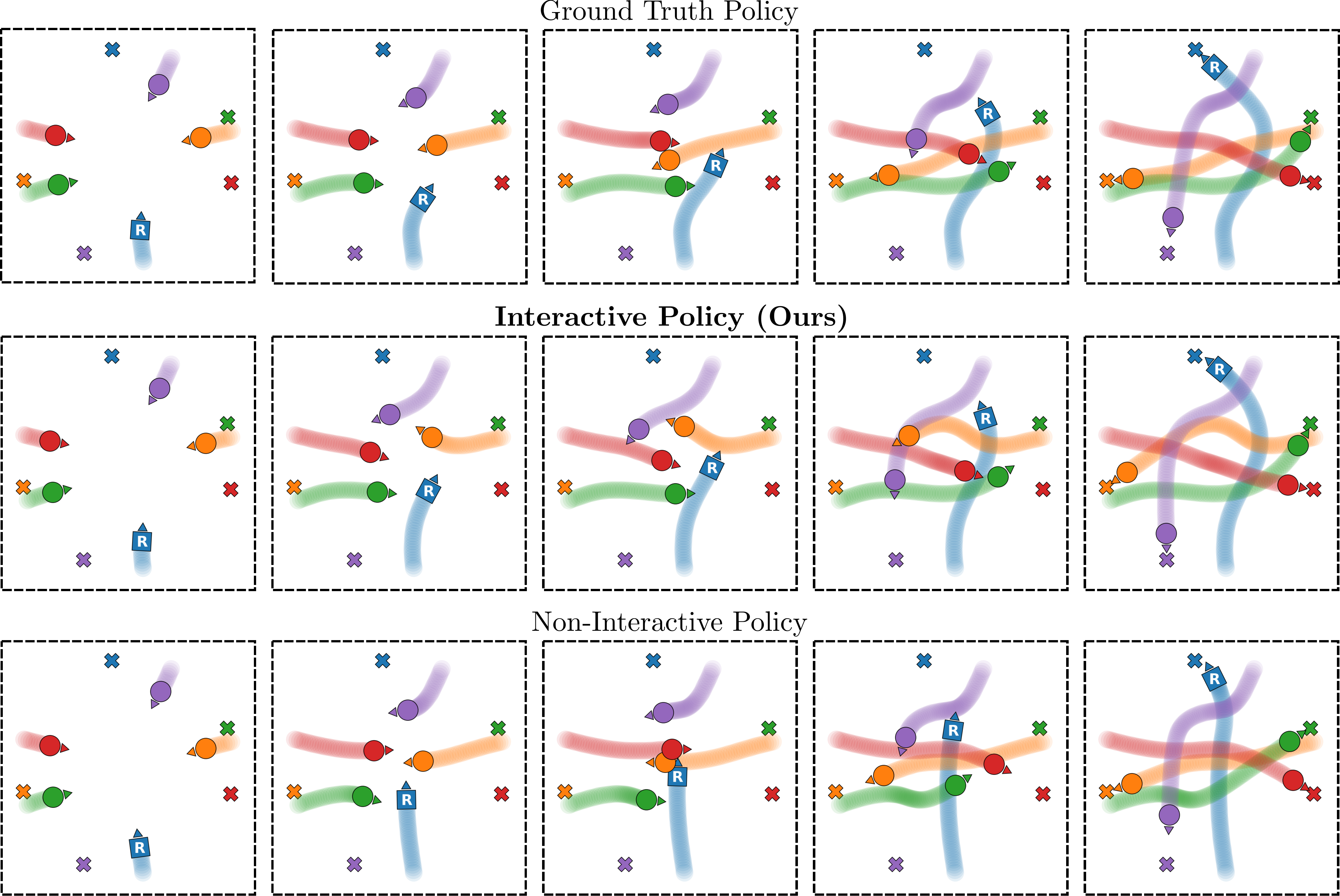}
    \caption{Qualitative results from the social navigation benchmark, where the letter ``R'' indicates the robot and the cross indicates the navigation goal of an agent. Learning from only 50 demonstrations, the proposed interactive policy significantly improves the safety performance of the non-interactive policy without compromising the efficiency, while performing comparably to the ground-truth policy.}
    \label{fig:snapshots}
    \vspace{-1em}
\end{figure*}

\begin{definition}[Non-interactive policy]
    A non-interactive policy of agent $j$, denoted as $\overline{\pi}_{\phi}^{(j)}(a|s)$, is an individual decision-making policy that describes the decision-making of agent $j$ without considering other agents. 
\end{definition}

Learning a non-interactive policy is equivalent to a standard single-agent imitation learning problem, where the multi-agent dataset is viewed as a collection of single-agent state-action sequences:
\begin{align}
    \phi^* = \argmax_{\phi} \sum_{i=1}^{N} \sum_{j=1}^{M_i} \sum_{t=1}^{T_i} \log \overline{\pi}_{\phi}^{(j)} (a_{i,j,t} \vert s_{i,j,t}). \label{eq:ind_mle}
\end{align} To simplify notation, we denote the non-interactive policy $\overline{\pi}_{\phi}^{(j)}(a|s)$ with the optimal parameter $\phi^*$ as $\overline{\pi}^{(j)}(a|s)$.

Learning the non-interactive policies in (\ref{eq:ind_mle}) is significantly easier compared to learning interactive policies in (\ref{eq:mle}) due to the significantly simplified conditional variable—the non-interactive policy only depends on a single agent's state instead of the joint states of all agents—and this is a well-studied problem with various well-performing methods, such as conditional variational autoencoders~\cite{co-reyes_self-consistent_2018}, diffusion models~\cite{chi_diffusion_2024} and flow-based models~\cite{chisari_learning_2024}.

\begin{definition}[Interaction game]
    Each agent in the interaction game optimizes an individual policy $\pi^{(j)}(a\vert s)$ with respect to an individual objective that depends on other agents' policies:
    \begin{align}
        J^{(j)}(\pi^{(1)}, \dots, \pi^{(M)}) = \sum_{k\neq j}^{M}\mathbb{E}_{\pi^{(j)}, \pi^{(k)}}[l_{\gamma}] + D_{KL}(\pi^{(j)} \Vert \overline{\pi}^{(j)}), \label{eq:game_obj}
    \end{align} where $l_{\gamma}(s, a, s^\prime, a^\prime)$ is a parameterized joint loss function for the state-action pairs from two policies $\pi^{(j)}(a|s)$ and $\pi^{(k)}(a^\prime|s^\prime)$, and $D_{KL}$ is the KL-divergence. 
\end{definition}

The first term in the objective function (\ref{eq:game_obj}) represents the collective intent shared among all the agents, such as avoiding collisions with others in navigation tasks. The second term in (\ref{eq:game_obj}) represents the individual intent of the agent, where the KL-divergence regulates the agent's current policy from deviating away from the agent's non-interactive policy.  

\begin{definition}[Nash equilibrium]
    A set of policies form a Nash equilibrium, denoted as $(\pi^{(1)^*}, \dots, \pi^{(M)^*})$, if and only if the following holds for all agents~\cite{nash_equilibrium_1950}:
    \begin{align}
        \pi^{(j)^*} = \argmin_{\pi^{(j)}} J^{(j)}(\pi^{(1)^*}, \ldots, \pi^{(j)}, \ldots, \pi^{(M)^*}), \text{ } \forall j. \label{eq:nash}
    \end{align}
\end{definition} 

The intuition behind Nash equilibrium is that it describes the scenario where all agents are simultaneously satisfied with its current policy given other agents' current policies, in which case no rational agent is willing to unilaterally change the policy.  

\begin{assumption}\label{assumption:nash}
    We assume the interactive policies in (\ref{eq:mle}) form a Nash equilibrium of the game formula (\ref{eq:game_obj}). 
\end{assumption}

This assumption integrates a game-theoretic structure into the formulation of the interactive policies in the MLE problem (\ref{eq:mle}), where we formulate the decision-making of each expert agent in the dataset as an explicit game-theoretic optimization problem defined in (\ref{eq:game_obj}). Importantly, if the joint cost function $l_{\gamma}$ is known in (\ref{eq:game_obj}), an iterative optimization algorithm is proposed in~\cite{muchen_sun_mixed_2024} to efficiently solve for the Nash equilibrium (\ref{eq:nash}) with guaranteed convergence. Therefore, given the non-interactive policies, the interactive policies as Nash equilibrium are parameterized by the parameter $\gamma$ in the joint loss function (\ref{eq:game_obj}):
\begin{align}
     & \pi_{\gamma}^{(1)}(a_{1} \vert s_{1\ldots M}), \ldots, \pi_{\gamma}^{(M)}(a_{M} \vert s_{1\ldots M}) \nonumber \\
     & = NE(l_{\gamma}, \overline{\pi}^{(1)}, \ldots, \overline{\pi}^{(M)}, s_{1\ldots M}), \label{eq:forward}
\end{align} where $NE$ denotes the algorithm from~\cite{muchen_sun_mixed_2024} solving for the Nash equilibrium. 

\begin{figure}[t!]
    \centering
    \begin{subfigure}{0.24\textwidth}
        \centering
        \includegraphics[width=\linewidth]{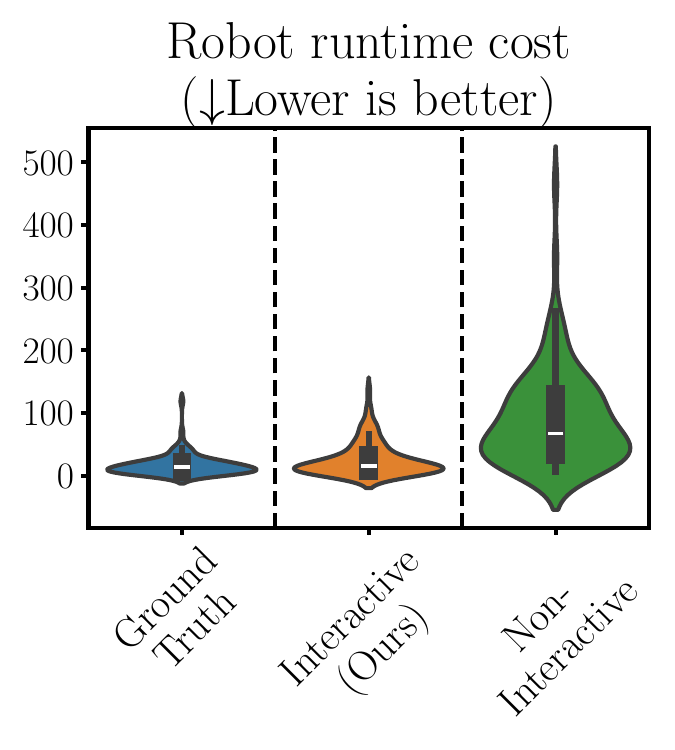} 
    \end{subfigure}
    \begin{subfigure}{0.24\textwidth}
        \centering
        \includegraphics[width=\linewidth]{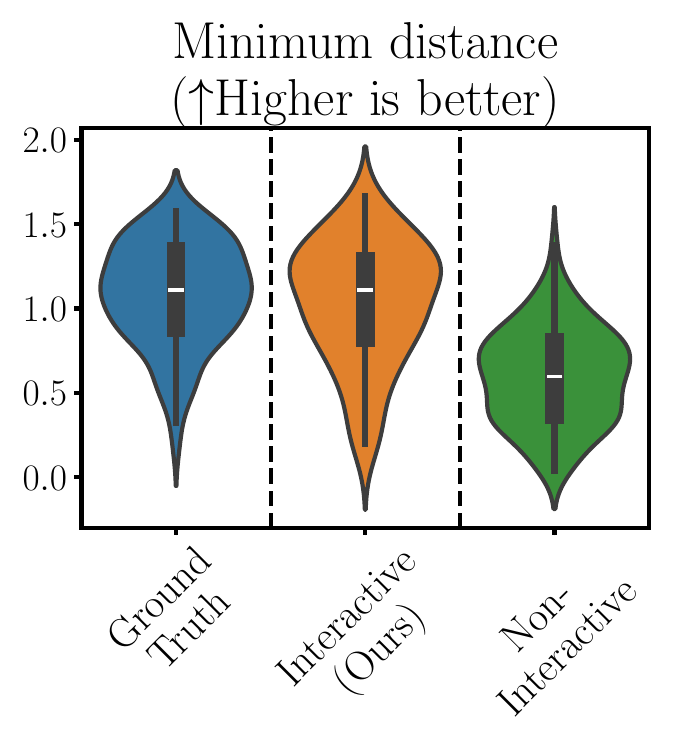}
    \end{subfigure}
    \vspace{-2em}
    \caption{Quantitative results of the social navigation benchmark (median, quartiles, and distribution of the metrics). The proposed interactive policy has comparable performance with the ground-truth policy and outperforms the non-interactive policy.}
    \vspace{-1em}
    \label{fig:results}
\end{figure}

\subsection{Learning interactive policies as inverse games}

Fully specifying the interactive policies as the Nash equilibrium of the game formula (\ref{eq:game_obj}) requires specifying the joint loss function $l_{\gamma}$, which is unknown a priori. In~\cite{sun_inverse_2025}, it is shown that the interactive policies (\ref{eq:forward}) under Assumption~\ref{assumption:nash} are differentiable with respect to the joint loss function parameter $\gamma$. Therefore, we can formulate the MLE problem for interactive policies (\ref{eq:mle}) as the following MLE problem for learning the joint loss function:
\begin{gather}
    \gamma^* = \argmax_{\gamma} \sum_{i=1}^{N} \sum_{j=1}^{M_i} \sum_{t=1}^{T_i} \log \pi_{\gamma}^{(j)}(a_{i,j,t} \vert s_{i,1\ldots M_i,t}), \label{eq:game_mle} \\
    \text{s.t. } \pi_{\gamma}^{(1)}, \ldots, \pi_{\gamma}^{(M_i)} = NE(l_{\gamma}, \overline{\pi}^{(1)}, \ldots, \overline{\pi}^{(M)}, s_{i,1\ldots M_i,t}), \text{ } \forall i. \nonumber
\end{gather} Since the calculation of the Nash equilibrium is differentiable, we model the joint cost function $l_{\gamma}$ as a neural network (e.g., a multi-layer perceptron) and solve the MLE problem (\ref{eq:game_mle}) through backpropagation. 

This problem of learning the cost function of a game formula is known as \emph{inverse games}, which is the multi-agent equivalent of the inverse optimal control (IOC) or inverse reinforcement learning (IRL) problem.

\section{Experiment}

\subsection{Experiment design}

We design a social navigation task with 100 randomized trials, where a group of 5 agents (one of them being the robot during tests) coordinate collision avoidance while reaching their individual goals. We simulate the agents using the iLQGames algorithm~\cite{fridovich-keil_efficient_2020}, a commonly used dynamic game solver. We model each agent as a circular disk under the Dubins car dynamics. The individual runtime cost function of each agent for iLQGames is specified by a navigation goal, a preferred longitudinal velocity, and a straight line reference trajectory from the current position to the goal with the preferred longitudinal velocity. 

\subsection{Implementation details}

We implement both the iLQGames algorithm and our method in JAX~\cite{bradbury_jax_2018} and Flax~\cite{heek_flax_2024}. We implement the non-interactive policy as a conditional variational autoencoder (CVAE). We implement the joint cost function as a multi-layer perceptron (MLP). We collect 50 navigation trials as the training data, where all 5 agents are simulated using iLQGames. In each trial, we uniformly sample the initial position of each agent on a circle and uniformly sample the parameters for the individual runtime cost function of each agent. 

\subsection{Experiment results}

We compare the proposed interactive policy with the ground truth iLQGames policy and the non-interactive policy that the Nash equilibrium is calculated based on. Qualitative results from one representative navigation trial are shown in Fig.~\ref{fig:snapshots}, where we show the actions of the robot controlled by different policies with the same condition. In each trial, the non-robot iLQGames agents operate under the assumption that the robot is an iLQGames agent with a presumed runtime cost function. To quantitatively evaluate how closely each policy behaves compared to the presumed iLQGames policy, we evaluate the state-action trajectories produced by each policy under the corresponding iLQGames runtime cost function, with the results shown in Fig.~\ref{fig:results} (left). Since navigation is a safety-critical task, we further evaluate the minimum distance between the robot and other agents in each trial, shown in Fig.~\ref{fig:results} (right).

From the qualitative and the quantitative results, we can see that the proposed interactive policy significantly outperforms the corresponding non-interactive policy. In particular, the interactive policy improves the safety performance (minimum distance) without compromising the navigation efficiency (runtime cost). Furthermore, we can also see that the proposed interactive policy performs comparably to the ground-truth iLQGames policy, closely imitating its behavior from only 50 demonstrations. These preliminary results demonstrate the potential of structured imitation learning methods in interactive environments with a limited number of demonstrations. 

\section{Conclusion and Discussion}

This work addresses the problem of imitation learning of interactive policies from multi-agent demonstrations. Despite the inherent high-complexity of the problem, we propose a structured imitation learning framework that combines single-agent generative model-based imitation learning with a game-theoretic interaction model. Through a social navigation benchmark, we show that leveraging explicit structure for modeling multi-agent interaction significantly improves the data efficiency, where the proposed interactive policy can perform comparably to the ground truth policy from only 50 demonstrations.  

The proposed method is compatible with any generative model-based single-agent imitation learning method, since the game-theoretic optimization problem (\ref{eq:game_obj}) can be solved using arbitrary non-interactive policies. Ongoing work includes integrating the method with a wider range of imitation learning methods and expanding the task beyond navigation to domains such as cooperative manipulation. Lastly, we plan to investigate the game formula (\ref{eq:game_obj}) further, aiming to improve the computational efficiency of the inverse game process, enabling rapid learning and adaptation of interactive policies with online observations. 

\section*{Acknowledgments}

This work is supported by the National Science Foundation grant CNS-2237576. The views expressed are the authors' and not necessarily those of the funders.

\bibliographystyle{plainnat_titlelink}
\bibliography{references}

\end{document}